\pdfoutput=1

\documentclass[11pt]{article}

\usepackage[]{acl}

\usepackage{times}
\usepackage{latexsym}
\usepackage{graphicx}
\usepackage[T1]{fontenc}

\usepackage[utf8]{inputenc}
\usepackage{multirow}
\usepackage{microtype}
\bibpunct[, ]{(}{)}{;}{a}{,}{,}
\title{TRBLLmaker - Transformer Reads Between Lyrics Lines maker}

\author{Mor Ventura \\
  Technion - IIT \\
    \texttt{\small{mor.ventura@campus.technion.ac.il}} \\
    \And Michael Toker \\ 
    Technion - IIT \\
    \texttt{\small{tok@campus.technion.ac.il}}}

\begin{document}
\maketitle

\begin{figure*}[t]
  \includegraphics[scale=0.48]{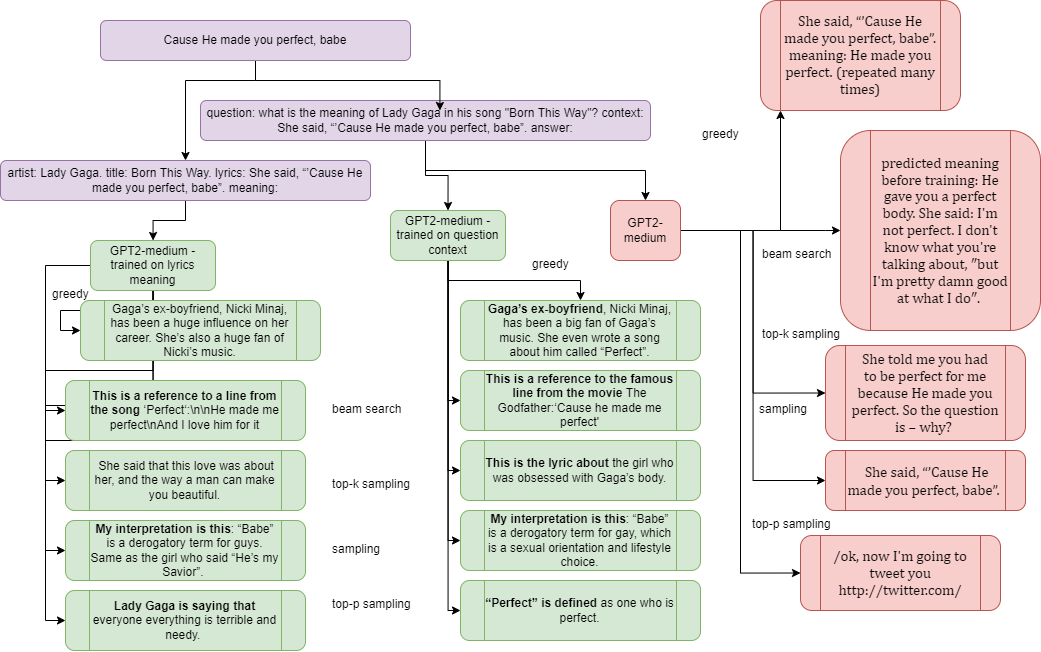}
\captionof{figure}{In this figure you can find a full example of prompts that we give the models as an input, and the output of the models with several decoding methods. We can notice that there is a huge improvement from the performance of the model before and after training.}
\label{fig: full_example}
\end{figure*}

\begin{abstract}
Even for us, it can be challenging to comprehend the meaning of songs. As part of this project, we explore the process of generating the meaning of songs. Despite the widespread use of text-to-text models, few attempts have been made to achieve a similar objective. Songs are primarily studied in the context of sentiment analysis. This involves identifying opinions and emotions in texts, evaluating them as positive or negative, and utilizing these evaluations to make music recommendations. In this paper, we present a generative model that offers implicit meanings for several lines of a song. Our model uses a decoder Transformer architecture GPT-2, where the input is the lyrics of a song.  Furthermore, we compared the performance of this architecture with that of the encoder-decoder Transformer architecture of the T5 model. We also examined the effect of different prompt types with the option of appending additional information, such as the name of the artist and the title of the song. Moreover, we tested different decoding methods with different training parameters and evaluated our results using ROUGE. In order to build our dataset, we utilized the 'Genious' API, which allowed us to acquire the lyrics of songs and their explanations, as well as their rich metadata.


\end{abstract}

\section{Introduction}
Transformers \citep{vaswani2017attention} are designed to handle sequence-to-sequence tasks and long-range dependencies effectively. There have been numerous attempts to generate reliable texts for tasks such as translation, summarization, dialog, and question answering. Many Transformer models were trained to perform tasks that required output that could be found in the input. These approaches are unsuitable for dealing with complex texts, hidden clues, or subtle implications.

As part of this paper, we address one of the most challenging generation tasks - reading between the lines of a song's lyrics. For an accurate interpretation of a text, and especially when explaining poems or any other textual art, we need to understand the environment, the context, and the semantics of the text. In this study, we examined the performance of the Transformer models, T5 \citep{JMLR:v21:20-074} and GPT2 \citep{radford2018improving}, with a variety of prompt types and decoding methods. According to the evaluation, the decoder Transformer only is more suitable for this task, prompts with meta data are useful, and the sampling decoding method produces better results. These conclusions provide us with more realistic meanings for lyrics as shown in the example in Figure \ref{fig: full_example}. A new dataset\footnote{MorVentura/TRBLLmaker} and API \footnote{tokeron/TRBLLmaker} are available in HuggingFace. 

\section{Related Work}
Among the core tasks of modern Natural Language Processing (NLP) applications is text generation. As opposed to many projects that concentrate on tasks such as machine translation, abstract summarization, or paraphrasing, we decided to construct a model to explain the meaning of song lyrics. To the best of our knowledge, this is the first study aimed at predicting the meaning of a song using a generative model. Abstractive summarization is the closest topic to our work.
\newline\newline
\noindent\textbf{Text Summerization}. Text summarization can be classified into two distinct families: extraction and abstraction. Using the extraction summarization method, key elements of the text are identified and transcribed verbatim. In contrast, abstractive summarization involves interpreting and analyzing the original text in order to produce a new, shorter text that emphasizes the most important points. This latter task is more challenging since it involves complex problems involving semantic representations and contexts \citep{gupta2010survey}. In view of the fact that our work is more closely related to abstractive summarization, we will focus on this area \citep{Jointly, Statistics}. Our work is similar to that published in \citep{music_info_thesis}, which uses a variety of unsupervised methods to produce a summary of song lyrics.
\newline\newline
\noindent
\textbf{Information Retrieval Of Songs}. Several studies have been conducted in the area of information retrieval from songs. The Million Song Dataset (MSD) project \citep{bertin2011million} is a collection of audio features and metadata of a million music tracks. The MSD focuses primarily on audio data, however, complementary datasets, musiXmatch and WASABI \citep{buffa2021wasabi}, associate MSD songs with their lyrics. The dataset we present is one step further, including annotations for song meanings.

\section{Data}

\textbf{Data Extraction}. To train a model to generate text, a large amount of data is required. We wanted to obtain annotated lyrics of songs based on their interpretations. In the absence of a suitable pre-existing online dataset suitable for our specific task, we decided to build our own custom database. We worked with 'Genius' website \footnote{\url{https://genius.com/}} which contains global collections of song lyrics and provides annotations and interpretations to song lyrics as well as additional music knowledge. We used the 'Genius' API and created a private client in order to retrieve relevant raw data from Genius servers. A  custom object named 'SongInfo' was created for each sample with the following attributes: title, genre, annotations, and meta information about the song. Metadata includes the name of the artist, the song id in the server, the lyrics, and statistics such as page views. To begin, we extracted the songs based on the most popular songs in each genre - pop, rap, rock, country, and R\&B.  Afterward, we created a pool of 150 artists \ref{appendix: artists} associated with a variety of music styles and periods and extracted a maximum of 100 samples from each. We merged all the data, without repetition, into one final dataset. Following a cleaning of non-English lyrics, we obtained our final corpus that consists of 8,808 different songs with over 60,630 samples, where each sample represents a specific sentence from the song's lyrics. For access to this diverse corpus, we created TRBLLDataset, which is based on HuggingFace Dataset and can be downloaded from there. \newline\newline
\noindent
\textbf{Data Exploration}. We performed data exploration as a preliminary step to training \ref{appendix:data_exploration}. As we can see in Figure \ref{fig:hist_genere} and \ref{fig:hist_singer}, the songs were taken from various genres such as rock, rap, pop and country, and various artists such as Metallica, The Beatles, Ed Sheeran and Drake.

\begin{figure}
    \centering
    \resizebox{0.95\columnwidth}{!}{\includegraphics{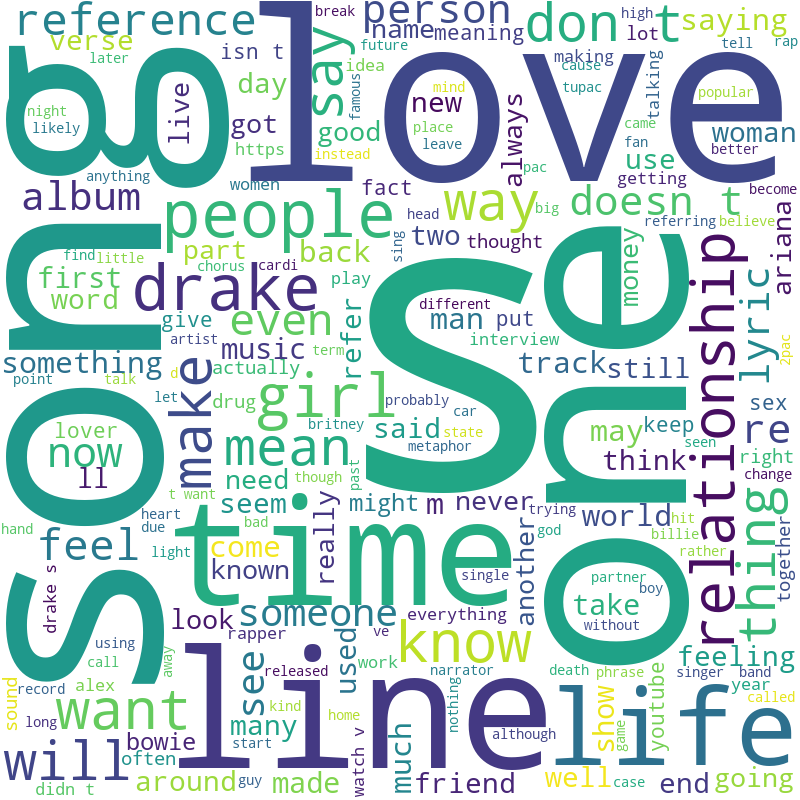}}
    \caption{wordcloud of words in annotations - the size of each word reflects the frequency of the words in the corpus.}
    \label{fig:wordcloud}
\end{figure}

Words-cloud which is based on annotations is shown in Figure \ref{fig:wordcloud} and words-cloud based on songs' lyrics is shown in Figure \ref{fig:wordcloud_lyrics}, while the size of word corresponds to the word frequency. Annotations commonly include words such as \emph{"song"}, \emph{"line"}, \emph{"reference"}, \emph{"one"}, \emph{"love"} and \emph{"relationship"}. After training, we expect the model to generate more of these words in its predictions. Words such as \emph{"nigga"}, \emph{"yea"}, \emph{"lala"}, \emph{"know"}, \emph{"want"}, and \emph{"love"} are frequently used in the lyrics. Here we can see the variety of words that come from different genres. It is interesting to observe whether the model can generalize and deal with different types of language and slang. Figures \ref{fig:hist_annot_len}, \ref{fig:boxplot_annot_len}, \ref{fig:boxplot_sample_len}, \ref{fig:hist_sample_len} illustrate the length of the samples and the annotations.
All songs with non-English symbols and their annotations have been deleted. Furthermore, we removed URLs from the annotations, and kept these examples without them. The data analysis revealed that some annotations refer to facts from the singer's personal life or to events that occurred close to the time the song was written. Many references are evident in the annotation of different related songs, personal events, artists, and the partners of the artists.  

\section{Models and Parameters}

\textbf{Models Architectures}. To begin with, we have attempted to use a full transformer as a generative model for our task. The model we used was T5-base. As part of our effort to better understand the performance of the model, we first made an inference, prior to fine-tuning, to determine how the model performed with its initial weights. The results were rather disappointing. Some of the sentences were in German, and in some cases, the output was empty. The model did not improve significantly after training for various parameters. Short training sessions or long training sessions with a low learning rate produced unintelligible results. The model generated empty strings after a long period of training, probably due to a bug. We decided to try a different architecture - a decoder only architecture. GPT2 and GPT2-medium were used. Additionally, we tested GPT-neo, which gave us excellent results even before training began. Since GPT-neo was too large for training on the given resources, only GPT2 and GPT2-medium were trained.
\newline\newline
\noindent
\textbf{Prompts}. As a result of GPT model rising, we have gained new insights regarding the way in which language models can be further enhanced by incorporating natural language prompts \citep{early_prompt}. Prompts are pieces of text inserted with input so that the original task can be transformed into a masked language modeling problem \citep{TianyuGao}. There are two different paradigms in the research of prompts - for small Language Model (LM) and for large LM. In small LMSs, fine tuning prompts is used to guide learners towards being better few-shot learners, while in super-large LMSs, different prompts are used to guide learners to perform different tasks since fine-tuning them requires modifying their parameters. Most of the work takes manually-designed prompts — prompt engineering is non-trivial since a slight perturbation can significantly affect the model’s performance, and creating a successful prompt requires both understanding of LMs' inner workings and trial-and-error. Here, we used three possible prompts with and without metadata, resulting 6 experiments in total:\textbf{lyrics meaning prompt} : "lyrics: [input] meaning: {annotation}", \textbf{task-with-metadata prompt} : "explain the song [title], written by [artist]. lyrics: [input]. meaning: [annotation]" and \textbf{question-context prompt} : "question: what is the meaning of [artist] in his song [title]? context: [lyrics]. answer: [annotation].\\

\begin{figure*}[t]
\includegraphics[scale=0.5]{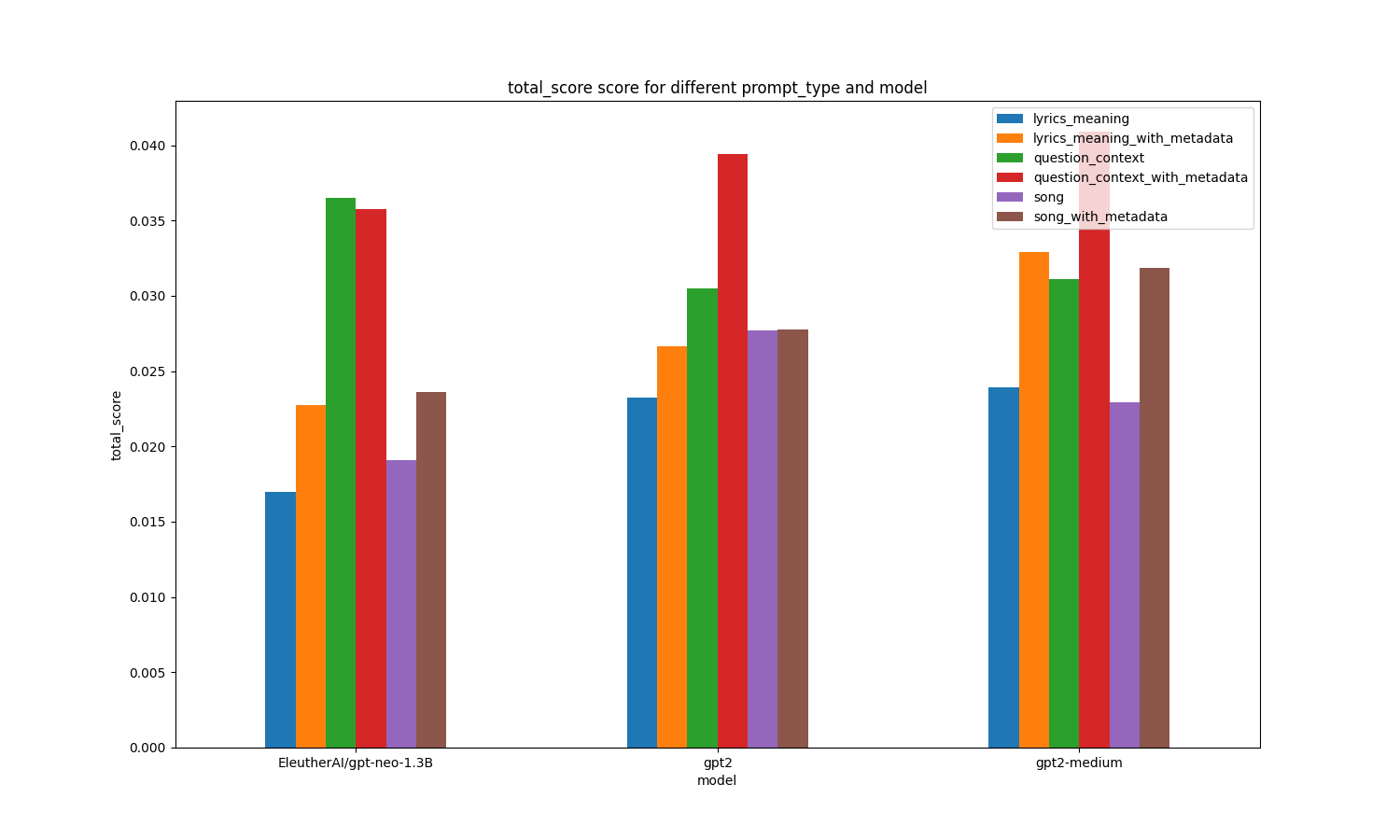}
\captionof{figure}{Total score for different prompt types and models}
\label{fig:score_diff_prompts}
\end{figure*}

\noindent
\textbf{Decode methods}. The quality of the sequence generated by the model depends on the decoding strategy used. There are advantages and disadvantages to each of these techniques.

\noindent
\underline{Greedy}: Greedy search simply selects the word with the highest probability as its next word.

\noindent
\underline{Beam search}: A beam search minimizes the risk of missing hidden high probability word sequences by keeping the most likely number of hypotheses at each time step and eventually selecting the hypothesis with the highest overall probability \citep{zarriess2021decoding}. Beam search will always produce a sequence of outputs with a higher probability than greedy search, but is not guaranteed to produce the most likely sequence of outputs. Even when the "no repeat ngram size" flag is used, the output still includes repetitions of the same word sequences. Using beam search is well suited to tasks where the length of the desired generation is more or less predictable, such as machine translation or summarization, but it is not well suited to open-ended tasks. The following parameters were defined: num beans = 3, no repeats, ngram size = 2, and early stopping.

\noindent
\underline{Sampling}: Sampling means randomly picking the next word 
according to its conditional probability distribution. It means the generation is not deterministic anymore. We defined: temperature: 0.95.

\noindent
\underline{Top-k sampling}: In Top-K sampling, the K most likely next words are filtered, and the probability mass is then redistributed among only those K next words \citep{fan2018hierarchical}. This sampling scheme was adopted by GPT2. As a result, our sampling pool is limited, which may result in the model producing gibberish for sharp distributions and limiting the model's creativity for flat distributions \cite{top_k_sampling}. We defined: top k = 50.

\noindent
\underline{Top-p sampling (nucleus)}: Top-p sampling chooses from the smallest possible set of words whose cumulative probability exceeds the probability p \citep{holtzman2019curious}. Top-p sampling picks the minimum number of words to exceed together \citep{top_p_samplin}. We defined: top p = 0.92.


\begin{figure*}[t]
\includegraphics[scale=0.7]{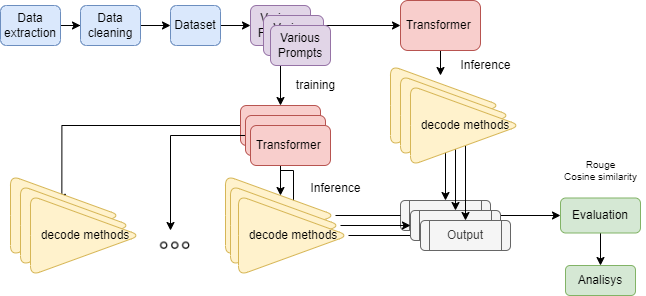}
\captionof{figure}{General flow of the research. After data extraction and cleanings, we generated various prompts and used them for evaluation transformers models and for fine tuning. We used various decoding methods.}
\end{figure*}

\section{Experiments and Results}

\textbf{Evaluation}
The evaluation metric for text generation models depends on the generation task - whether we translate, summarize, answer to a question or generate new information that is not mentioned explicitly in the input, such as required for our case. We conducted two phases of evaluation: The first one was performed before any training and the second after it. Different models with different prompts and different decode methods were compared: 
Three candidate models: EleutherAI/gpt-neo-1.3B, gpt2-medium and gpt2, 
with seven prompts types: 'lyrics - meaning', 'question - context', 'song', each with optional additional metadata of artist and song's title and without prompt - 'None',
and five decode methods: beam search, greedy, sampling, top-k sampling an top-p sampling.
We computed two main types of metrics, ROUGE and cosine similarity. So they were computed based on 105 combinations over ten fixed samples:

\noindent
\underline{ROUGE-1}: The ROUGE-N metric measures the number of matching 'n-grams' between our model generated meaning and the ground truth annotation \citep{lin2004rouge}. Because of the complexity of our task, we decided to work with unigrams (N = 1), which means the generated text is not expected to consist of consecutive words found in the training input.

\noindent
\underline{Cosine similarity}: An inner product space measuring the cosine of the angle between two non-zero vectors is referred to as a cosine similarity. Following the creation of a bag of words, we created vectors of counts of the number of times each word appears in each text. Based on those vectors, we calculated the cosine similarity. By this metric, we decided how close our prediction was to the ground truth and how close it was to the song's lyrics \citep{li2013distance}.

\noindent
\underline{Total score}: We created an integrated score based on ROUGE-1, cosine-similarity p-a (prediction - annotation) and cosine similarity p-l (prediction - lyrics):

\[Total Score = \sum_{(i) \in M}{\alpha_{i}\cdot m_{i}} \]
\[Total Score = \alpha_{1}\cdot Rouge + \alpha_{2}\cdot CS_{p-a} - \alpha_{3}\cdot CS_{p-l}\]
\[Total Score = \max(0, Total Score)\]

Where \(m\) stands for metric and \(alpha\) for weight.
The weight was set to 0.5. We penalize a generated meaning that has high cosine similarity with the lyrics. The logic behind it is to get indication to some kind of memorization of the lyrics or meaningless repetitions of the input. We normalized the metric , So our metric total score spans between [0,1].

In the first phase, we examined all the candidate models and parameters. We used it to help us conclude which of the candidates is the best starting point to train on. In the last phase we, examined the trained models over the best prompts chosen in the first phase.

\subsection{Performance}

\textbf{Parameters}

\noindent
\underline{Batch size}: Since we were limited in resources (GPU), the maximal batch size we could get is 2. This was too small batch size and resulted in unstable training. We used gradient accumulation in order to simulate larger batch size. We tried 16, 32 and 64 and got the best results with models trained with batch size grater than 32.

\noindent
\underline{Train epochs}: We notice that the results were not improving after more than 8 epochs.

\noindent
\underline{Learning rate}: The best learning rate achieved with 0.000005

\noindent
\textbf{Prompts}

\noindent
First, we wanted to examine whether there was a difference in performance between models trained on different prompts. The second aim of this study was to determine whether the performance of prompts that contain song metadata (song name and singer name) has improved. A higher score was achieved by prompts with metadata as compared to prompts without metadata, as shown in Figure \ref{fig:score_diff_prompts}. This result is consistent with our intuition, since many annotations use the name of the artist and title of the song. Further, the model learned that there is a relationship between artist names, and refers to related singers. For example, when explaining the song \emph{"Born This Way"} by \emph{"Lady Gaga"}, the generated meaning was - \emph{"Gaga’s ex-boyfriend, Nicki Minaj, has been a big fan of Gaga’s music"}.

Figure \ref{fig:score_diff_prompts} illustrates the best score achieved by 'question context with metadata' and 'lyrics meaning with metadata'. As these prompts are similar to how GPT2 was trained, the model is capable of learning quickly and generating suitable annotations. The scores of different decoding methods and prompts are presented in Table \ref{table:dec_scores} and \ref{table:dec_scores_after_training}.

\noindent
\textbf{Decode methods}

As can be seen from the comparison of these methods, there are quite significant differences between them. \underline{Greedy} search has the major disadvantage of missing high-probability words hidden behind a low-probability word, and it performs poorly in most cases. As we have seen, \underline{beam search} suffers from repetitive generation, particularly before training, although it does produce a more realistic and relevant generated text in relation to the metadata provided in the input. When adapting the model's training objective, beam search can generate more fluent text than Top-P sampling \cite{beam_search_better}. The model learned structures that characterize how people explain lyrics. In some cases, it generates the meaning based on the name of the artist (\emph{"Cole is saying that he .."}) or by referring to specific words in the input (\emph{"This is a reference to the..."}, as demonstrated in the left side in bold of Figure \ref{fig: full_example}. \underline{Sampling}, \underline{top-k sampling}, and \underline{top-p sampling} seem to produce more fluent texts. As humans, we like it when generated text surprises us and is not monotonous or predictable. As a result, we observe more realistic meaning with connections to additional references that are associated with the artist or title of the song. Naturally, this adds false information most of the time and makes up fake facts, for example, the following generated meaning\emph{"Gaga has been a pop icon since the late 1950s..."} is nonfactual, since \emph{Gaga} had her breakthrough in the 2000s. In comparison to other sampling methods, top-p sampling appears to produce a more complex and meaningful result.(See the \ref{appendix: examples} for more fun examples)
The methods were compared before and after training.
Among the three metrics, top-p sampling, top-k sampling, and beam search had the best performance. In comparison to the pretrained model, we obtained a 37 percent increase in the total score after training, as shown in Tables \ref{table:dec_scores} and \ref{table:dec_scores_after_training}.\\\\\\

\begin{table*}[t]
\label{tab:table2}
  \begin{tabular}{|l|c|c|c|c|c|c|}
    \hline
    {Decode Methods} &
      \multicolumn{2}{c}{Total score} &
      \multicolumn{2}{c}{Rouge-1} &
      \multicolumn{2}{c}{Cosine Similarity} \\
    & lyrics-meaning (lm) & question-context (qc) &lm & qc &lm & qc \\
    \hline
    beam search & 0.40  & 0.56 & 0.038  & 0.039  & 0.19  & 0.20  \\
    \hline
    greedy & 0.21  & 0.25  & 0.037  & 0.038   & 0.19  & 0.17   \\
    \hline
    sampling & 0.30   & 0.33  & 0.037  & 0.037  & 0.14   & 0.15  \\
    \hline
    top-k sampling & 0.30  & 0.48   & 0.038  & 0.47   & 0.18   & 0.17  \\
    \hline
    top-p sampling & 0.42  & 0.41   & 0.035   & 0.036   & 0.15   & 0.18   \\
    \hline
  \end{tabular}
  \caption{
  different decoding methods scores - before training.\\ 
  The following table shows the scores according to the metrics we calculated.
  \\
  between two prompt types: lyrics-meaning prompt and question-context prompt with meta data (gpt-2 medium)}
  \label{table:dec_scores}
\end{table*}

\begin{table*}[t]
  \begin{center}
    \label{tab:table1}
    \begin{tabular}{l|l|l|l|l}
      \textbf{Model} & \textbf{decode} & \textbf{rouge1} & \textbf{cosine}   & \textbf{total}\\ 
      \hline
      lyrics meaning & beam search & 0.038 & 0.21 & 0.55 \\ 
                & top k & 0.036 & 0.24  &  0.39 \\ 
                & top p & 0.042 &  0.2 & 0.5 \\
      question context & beam search & 0.042 & 0.23 & 0.65 \\
                & top k & 0.04 &  0.24 & 0.48 \\
                & top p & 0.036 & 0.2  &  0.47 \\
    \end{tabular}
      \caption{different decoding methods scores - after training.\\}
      \label{table:dec_scores_after_training}
  \end{center}

\end{table*}

\section{Further Discussion and Conclusions}

This paper presents the performance of the T5 and GPT2 Transformer models on a complicated generation task - reading between the lines of song lyrics. In other words, understanding the implicit meaning of poetic expressions. The results of T5 were poor, and it was rather fixated on the translation task, which was one of its training exercises. In contrast, GPT-2 demonstrated better results and sometimes entertained compliance with our requirements.We explored the field of sequence to sequence generation, and demonstrated that the wording of the input has a significant impact. The way we present the task for the model is critical. Prompts such as question-context improved the quality of the generated text. The decode method also played an important role, sampling methods outperformed all the rest.

Several components need to be examined. First, due to the lack of appropriate computational resources, we would like to train also GPT-neo, which requires more memory. As a second step, we must continue to evaluate the similarity between the generated meaning and the ground truth using a metric that emphasizes the similarity in meaning, rather than only the similarity in words. The 'SiameseLSTM' metric \citep{SiameseLSTM}, LDA for topics similarity \citep{yu2001direct}, and the Word Mover distance \citep{huang2016supervised} are examples of this type of analysis. Additionally, we plan to train the model in a modular manner, which means that we will update the weights continuously with a number of prompts.
We intend to investigate the results using explainability techniques in order to gain a better understanding of the model implication process and to derive suitable conclusions. Moreover, we would like to demonstrate compatibility between any given text and any given song. Our future plans include applying our model capabilities to additional advanced applications, such as fitting songs to any free text based on their mutual meaning. The application of attention-based models to tasks such as meaning generation is something we are eager to explore in the future. 
Hopefully, this preliminary study will encourage more NLP researchers to improve the way we understand and enjoy songs since achieving artistic comprehension is another step towards achieving robust artificial intelligence.


\bibliography{acl_latex}
\bibliographystyle{acl_natbib}



\newpage
\section{Appendices}

\appendix
\section{Data exploration appendix}
\label{appendix:data_exploration}

\begin{figure}[!b]
    \centering
    \resizebox{0.95\columnwidth}{!}{\includegraphics{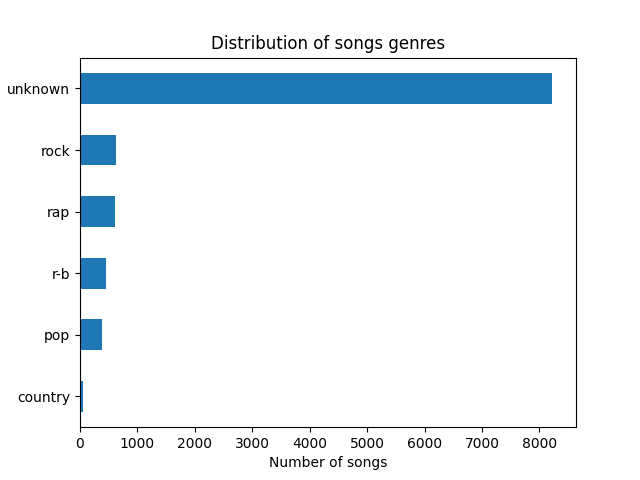}}
    \caption{Distribution of songs per genre.}
    \label{fig:hist_genere}
\end{figure}


\begin{figure}[!b]
    \centering
    \resizebox{0.95\columnwidth}{!}{\includegraphics{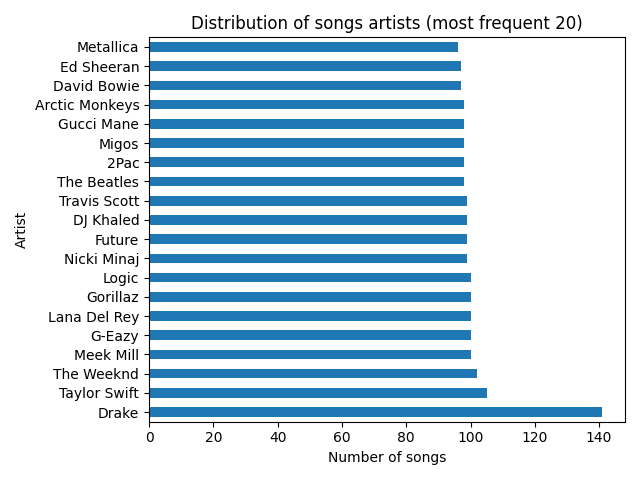}}
    \caption{Distribution of songs per artist.}
    \label{fig:hist_singer}
\end{figure}


\begin{figure}[!b]
    \centering
    \resizebox{0.95\columnwidth}{!}{\includegraphics{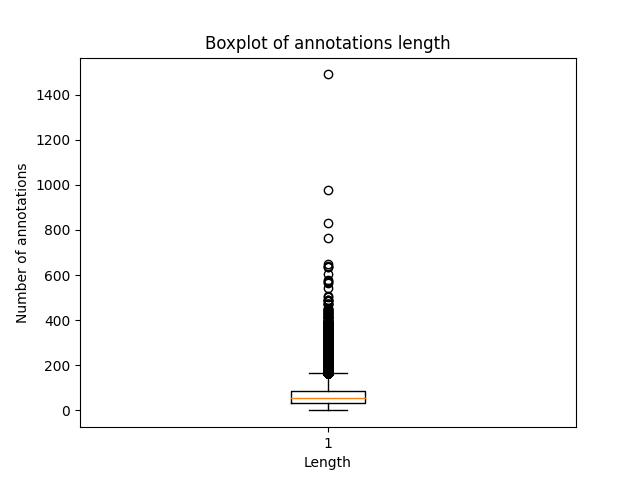}}
    \caption{Annotation length distribution.}
    \label{fig:boxplot_annot_len}
\end{figure}


\begin{figure}[!b]
    \centering
    \resizebox{0.95\columnwidth}{!}{\includegraphics{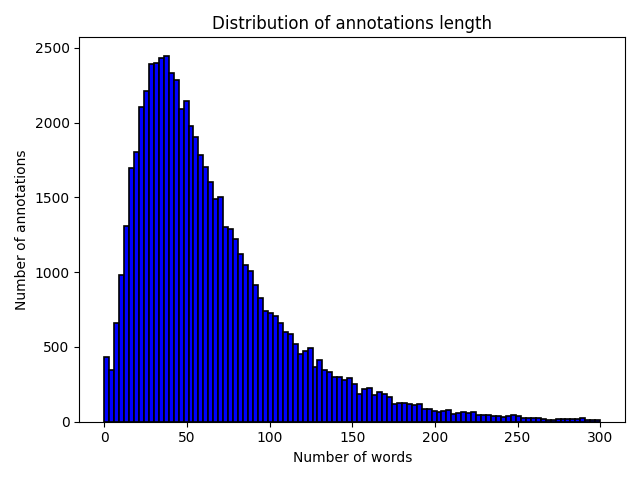}}
    \caption{\textbf{Annotation length histogram.}}
    \label{fig:hist_annot_len}
\end{figure}


\begin{figure}[!b]
    \centering
    \resizebox{0.95\columnwidth}{!}{\includegraphics{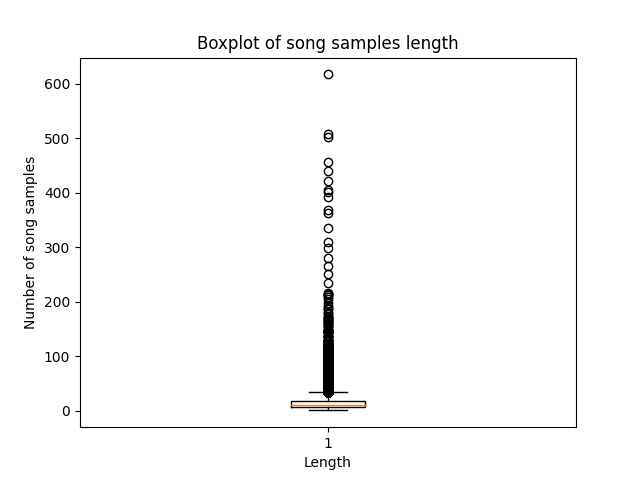}}
    \caption{\textbf{Samples length distribution.}}
    \label{fig:boxplot_sample_len}
\end{figure}


\begin{figure}[!b]
    \centering
    \resizebox{0.95\columnwidth}{!}{\includegraphics{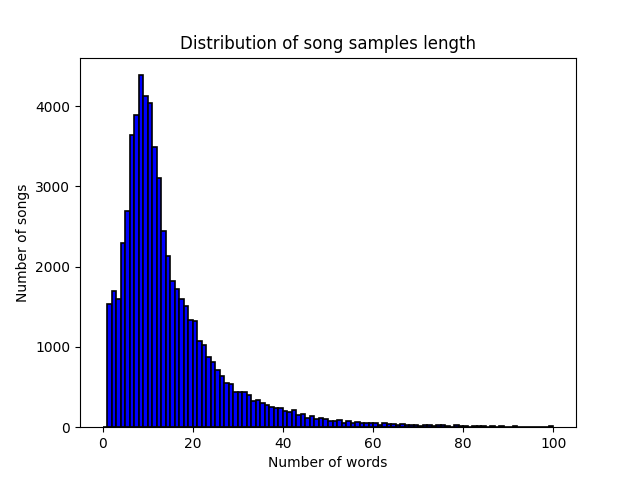}}
    \caption{\textbf{Samples length histogram.}}
    \label{fig:hist_sample_len}
\end{figure}


\begin{figure}
    \centering
    \resizebox{0.75\columnwidth}{!}{\includegraphics{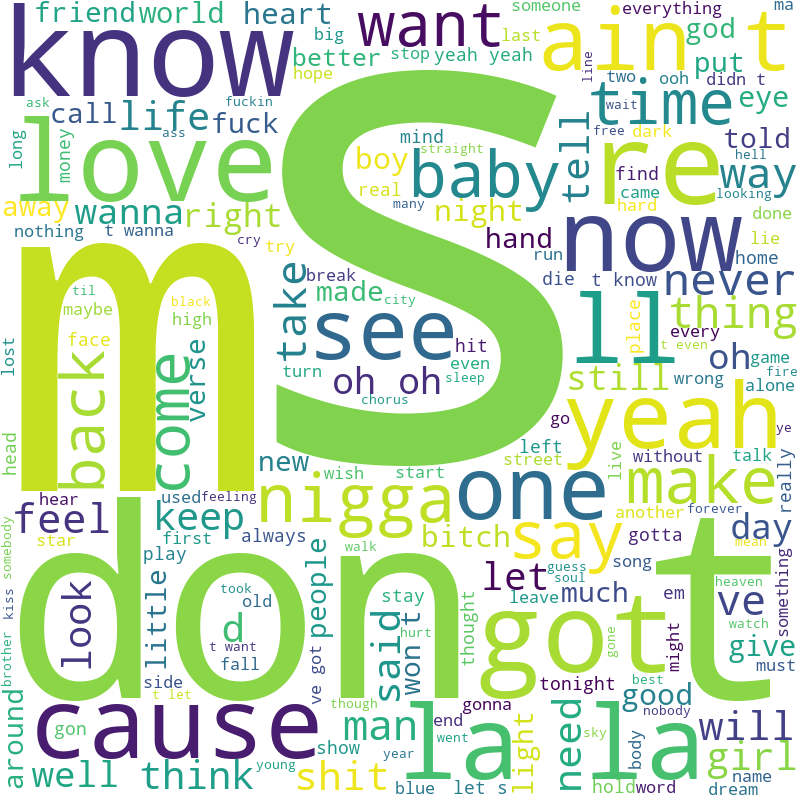}}
    \caption{\textbf{Wordcloud} of words in lyrics - the size of each word corresponds to the frequency of the words in the corpus.}
    \label{fig:wordcloud_lyrics}
\end{figure}


\newpage
~\newpage
\section{Examples Appendix}
\label{appendix: examples}
Several examples of meanings that were generated by gpt-2 medium with lyrics-meaning prompt and question-context prompt with different decode methods and additional meta data (It might be pretty funny!):\\
\color{red!60!black}Red\color{black} - for data that relies on lyrics and title.\\
\color{blue!60!black}Blue\color{black} - for data that relies on the artists names.\\
\color{green!60!black}Green\color{black} - general phrases the model learned for interpretations.\\
-----------------------------------------------------------\\
\noindent
1)
Song's title: \textbf{\color{red!60!black} Magic Shop\color{black} (English Translation)}\\
\newline\newline
\noindent
artist: \color{blue!60!black} BTS \color{black}(Korean artist)\\
prompt:lyrics-meaning\\
input: \textit{lyrics: \color{red!60!black} I wanted to become your comfort and move your heart. I want to take away your sadness, and pain\color{black}}. \\
\newline\newline
\noindent
\underline{greedy}:\\
 \textit{"\color{green!60!black}The first line is a reference to the song\color{blue!60!black} “Magic Shop” \color{black} by the Japanese band, \color{blue!60!black}OneRepublic\color{black}. \color{green!60!black}The song is about a \color{black} girl who is \color{red!60!black}depressed and lonely\color{black}. She wants to be comforted by a man who is able to \color{red!60!black}help\color{black} her."}
\newline\newline
\noindent
\underline{beam search}:\\
 \textit{"\color{green!60!black}This is a reference to the song \color{red!60!black}“Magic Shop” \color{black} by the Japanese pop group AKB48."} 
\newline\newline
\noindent
\underline{top-k sampling}:\\
 \textit{"This girl’s a type of \color{red!60!black}“comfort”\color{black} girl as well. She wants to \color{red!60!black}comfort \color{black} his person or body because she already has enough  comfort, but he is still \color{red!60!black}hurt \color{black} that he had to do something like this."} 
\newline\newline
\noindent
\underline{sampling}:\\
 \textit{"\color{green!60!black}Compared to popular belief\color{black}, an easy life with the comfort of a \color{red!60!black}lover\color{black} stands a different challenge for a man. \color{green!60!black}This figure is symbolical for a\color{black} woman trying to leave her \color{red!60!black}dull\color{black} life with her significant other."} 
\newline\newline
\noindent
\underline{top-p sampling}:\\
\textit{"After being pursued by the fans, or contractually given it by \color{red!60!black}BTS \color{black}, I wanted to release this track to you, to fill your lives with \color{red!60!black}beauty, joy, happiness\color{black}, etc. I want to change your relationship with anyone, anytime, anywhere."} 

-----------------------------------------------------\\
2)
Song's title: \textbf{\color{red!60!black}January 28th\color{black}}\\
artist: \color{blue!60!black} J. Cole \color{black}\\
prompt:lyrics-meaning\\
input:\\
\textit{artist: J. Cole. title: January 28th. lyrics: \color{red!60!black}This for my niggas that was tossed in the graves, Every so often I fade deep in my thoughts, And then get lost in the days, We used to play before your coffin was made, Just got the call nigga got caught with a stray, Hope he’s okay.\color{black} meaning:}
\newline\newline
\noindent
\underline{greedy}:\\
 \textit{"\color{blue!60!black}Cole \color{black}is saying that he’s not the only one who has been in a relationship with someone who was thrown in the \color{red!60!black}grave\color{black}. \color{green!60!black}He’s also saying that \color{black} he’s not the only one who has been in a relationship with someone who was thrown in the grave.\color{blue!60!black} Cole’s \color{black}“we used to play before your coffin was made” \color{green!60!black} could also be a reference to the fact that \color{black} he’s been in a relationship with someone who was thrown in the grave."} 
\newline\newline
\noindent
\underline{beam search}:\\
\textit{"\color{blue!60!black}Cole \color{green!60!black} is saying that he \color{black} has been through a lot in his life, but he still has hope that things will get better.\color{red!60!black}“Tossed in graves”\color{green!60!black} is a reference to the fact that \color{blue!60!black}Cole \color{black} was shot and killed by a \color{red!60!black}police\color{black} officer."} 
\newline\newline
\noindent
\underline{top-k sampling}:\\
\textit{"In 2009, the Uptown \color{blue!60!black}J. Cole \color{black}opened up about “Falling In Reverse”, one of his favorite songs. \color{green!60!black} Though he didn’t really know the subject at the time, this song influenced him for this\color{black} title: in the studio with Future, this song was the most played of \color{blue!60!black}Cole’s \color{black}songs. \color{blue!60!black}Cole’s \color{black}favorite song on the album was “Jumpin’, I’ve Been Goin’ Far,” \color{green!60!black} which is dedicated to\color{black} 
 Future.\color{green!60!black}”This is another example of a metaphorically twisted way..\color{black}"}

\noindent
\underline{sampling}:\\
\textit{"\color{green!60!black} The placement of\color{red!60!black} “throwin' in the graves” \color{green!60!black} reads like someone is saying that\color{black} \color{blue!60!black}Cole’s \color{black} friends have been killed by this person.\color{red!60!black}“Throwin' in the graves”\color{black} is also the first job title \color{blue!60!black}Cole\color{black} describes in \color{blue!60!black}El-P’s\color{black} “xxx2”.A miscarriage, \color{green!60!black}he is often seen on the set of\color{black}  Graduation: Dancing in the Cruises to remind his \color{red!60!black}friends\color{black} that someday (Whatever it is) they would be great at dozens of chicks (dancing in the Cruises, not a new subject in.."}

\noindent
\underline{top-p sampling}:\\
\textit{"\color{blue!60!black}Tyler \color{black} references the “\color{blue!60!black}J. Cole \color{black} Stranded Barretting” \color{green!60!black}he wrote about his\color{black}  freestyle on June 7, 2013” Spark Basket with \color{blue!60!black}Drake\color{black} coming out of his father’s \color{red!60!black}grave\color{black}, me and \color{blue!60!black}Coldplay’ll \color{black}get you out, \color{blue!60!black}Timbaland \color{black}with the mic, Jax made your night, Matthew McConaughey, Joyner Mims. It was just love at first sight "}

-----------------------------------------------------\\
3)
Song's title: \textbf{\color{red!60!black}Where the Streets Have No Name\color{black}}\\
artist: \color{blue!60!black}U2\color{black}\\
prompt:lyrics-meaning\\
input: \textit{artist: U2. title: Where the Streets Have No Name. lyrics: \color{red!60!black}We’re still building and burning down love, Burning down love, And when I go there, I go there with you, It’s all I can do\color{black}. meaning:}\\

\noindent
\underline{greedy}:\\
\textit{"\color{green!60!black} The song is about the \color{black} \color{red!60!black}destruction of love\color{black} and the destruction of the world. The song is about the destruction of love and the destruction of the world.The song is about the destruction of love and the destruction of the world.The song is about the destruction of love and the destruction of the world.The song is about the destruction of love and the destruction of the world.The song is about the destruction of love and the destruction of the world.The song is about the destruction of love and the destruction of the world.The.."}

\noindent
\underline{beam search}:\\
 \textit{"\color{green!60!black}This is a reference to the song \color{black} \color{red!60!black}“Burning Down\color{black} the House” by \color{blue!60!black}The Rolling Stones\color{black}. \color{green!60!black}The song is about a man who \color{black} has lost his wife and is trying to rebuild his \color{red!60!black}relationship with his lover\color{black}."} 

\noindent
\underline{top-k sampling}:\\
\textit{"In the song, \color{blue!60!black}U2 \color{black} sings about \color{red!60!black}loving someone\color{black}, \color{green!60!black}but not necessarily in the usual emotional context; as in,\color{black} “It’s all I can do.” \color{green!60!black} He states himself as being somewhat \color{black} introverted and “a bit quiet,” \color{green!60!black} but these lyrics are almost more about being the speaker of this song. The lyric and the delivery, as stated, make it sound like \color{blue!60!black}U2’s \color{red!60!black}love\color{black} for his subject has gone sour, but this can be interpreted as he is still \color{red!60!black}longing\color{black} and wants more, but he doesn’t even know where.."}
\newline\newline
\noindent
\underline{sampling}:\\
\textit{"\color{red!60!black}“Burning down love”\color{green!60!black} can be understood as one’s\color{black} awakening to one’s true (moral) nature. The “religious” part of \color{green!60!black} this phrase is accentuating the fact that\color{black}, as people come to terms with their own true nature, they’ll grow to fulfill their divine purpose of existence and their lives.Secondly, the power that Christians hold over humanity is a powerful force. By breaking away from the orthodox way of doing things and worshiping Jesus, they simply achieved a higher form of consciousness that is inaccessible to the establishment"}
\newline\newline
\noindent
\underline{top-p sampling}:\\
\textit{"\color{green!60!black}While the rest of the album is about \color{red!60!black}the effects of love\color{black}, the instrumental is the one that ends this section."} 

-----------------------------------------------------\\
4)
Song's title: \textbf{\color{red!60!black}G.U.Y\color{black}}:\\
artist: \color{blue!60!black}Lady Gaga \color{black}\\
prompt: question context\\
input: \textit{question: what is the meaning of Lady Gaga in his song "G.U.Y."? context: \color{red!60!black}I’m gonna wear the tie, want the power to leave you, I’m aimin’ for full control of this love (Of this love)\color{black}. answer:}

\noindent
\underline{greedy}:\\
\textit{"\color{blue!60!black}Gaga’s \color{black} \color{red!60!black}“love”\color{green!60!black} is a metaphor for \color{black} her \color{red!60!black}power\color{black}, and her \color{red!60!black}ability to control her lover’s\color{black} actions."}
\newline\newline
\noindent
\underline{beam search}:\\
\textit{"\color{green!60!black}This is a reference to the song\color{red!60!black} “Power to Leave You”\color{black} by \color{blue!60!black}The Chainsmokers\color{black}.https://www.youtube.com/watch?v=qZjx0qH7x8E"} 

\noindent
\underline{top-k sampling}:\\
\textit{\color{blue!60!black} Gaga’s \color{black} been a pop icon since the late 1950s, but she was so caught in that spotlight that her life seemed to be moving forward at full blast. She is trying to keep it all together, using her \color{red!60!black}power\color{black}.\color{green!60!black} In short, \color{black} he’s \color{red!60!black}been in love\color{black} with her too long. \color{green!60!black} She claims to\color{black} have made it her business, and if he has it, then she wants a chunk of it.\color{green!60!black}In the same vein of the previous line, \color{blue!60!black}Gaga \color{green!60!black} also said she thought of the song\color{black}}

\noindent
\underline{sampling}:\\
\textit{\color{green!60!black}This line could be referencing\color{blue!60!black} Lady Gaga’s\color{black} 2012 mixtape by \color{blue!60!black}Doja Cat\color{black}, “Luda”. The X is the first character \color{green!60!black}on the album, a reference to\color{black} knee-high stockings."} 
\newline\newline
\noindent
\underline{top-p sampling}:\\
\textit{\color{green!60!black} In a way, these lines exemplify\color{black} the “I’m all mine, ain’t nobody gettin’ in my way mindset in the recent economic crisis in America. \color{blue!60!black} Gaga \color{black}is pulling out all the stops in order to make sure she gets her way. \color{green!60!black}She’s likely going to paint a very striking image of how she is becoming\color{black} a full owner of this world of yours.She also makes a swipe at her Target knockoff, Belk Air “Snowbird” \color{green!60!black}which similarly\color{black} involves..}

-----------------------------------------------------\\
5)
Song's title: \textbf{\color{red!60!black}Stronger\color{black}}\\
artist: \color{blue!60!black}Britney Spears\color{black}\\
prompt: question context\\
input:\textit{question: what is the meaning of Britney Spears in his song "Stronger"? context:\color{red!60!black} Here I go, on my own, I don’t need nobody, better off alone, Here I go, on my own now, I don’t need nobody, not anybody, Here I go (Here I go, here I go, here I go...)\color{black}. answer:}
\newline\newline
\noindent
\underline{greedy}:\\
 \textit{\color{blue!60!black} Britney \color{black} \color{green!60!black} is saying that \color{black} \color{red!60!black}she’s not afraid\color{black} to \color{red!60!black}go her own way\color{black} and be herself. She’s not afraid to be herself and be herself. She’s not afraid to be herself and be herself. She’s not afraid to be herself and be herself. She’s not afraid to be herself and be herself. She’s not afraid to be herself and be herself. She’s not afraid to be herself and be herself. She’s not afraid to be herself and be herself.}
\newline\newline
\noindent
\underline{beam search}:\\
\textit{"\color{green!60!black}This line is a reference to the song\color{black} “I” by \color{blue!60!black}The Lumineers\color{black}."}
\newline\newline
\noindent
\underline{top-k sampling}:\\
\textit{\color{blue!60!black} Britney’s \color{black} \color{red!60!black}never needed anybody\color{black}, but having to have somebody watch her, and make sure she don’t fall off a cliff, is too tough.\color{blue!60!black}“Better off alone” \color{green!60!black} is a nice phrase from Britney Spears' early “Thicker Than Water”\color{black}. It expresses \color{blue!60!black} Britney’s\color{black} \color{red!60!black}dissatisfaction with her relationship\color{black} as her relationship with her previous husband (Brian Johnson). She now wants people to love her for herself, and to be her own person.Britney’s not ready}
\newline\newline
\noindent
\underline{sampling}:\\
\textit{Britney \color{green!60!black}is saying that \color{black} 
she’s happy and focused as she makes her way through the world. \color{green!60!black}But now that she’s adjusting to the fact that \color{black} she isn’t in the relationship, \color{red!60!black}she doesn’t need anyone to help her solve her problems\color{black}. \color{green!60!black}That being said,\color{black} there are so many networks that \color{blue!60!black}Britney\color{black} needs to control which makes it hard to get away from problems. Can’t hold it, keeps on hitting me straight in the face And I just have to russ that at Her phone}
\newline\newline
\underline{top-p sampling}:\\
\textit{\color{green!60!black}She felt \color{black}she wanted to make love and isn’t making any plans \color{red!60!black}to leave\color{black} her partner. So she stays.} 

\newpage
\section{List of artists Appendix}
\label{appendix: artists}

                 \textit{'Backstreet Boys', 'Spice Girls', 'Soulja Boy', 'ZAYN',
                'Travis Scott', 'Sam Smith', 'DJ Khaled', 'Pharrell Williams',
                'Stone Temple Pilots', 'Clean Bandit', 'Taylor Swift'
                '2Pac', 'Colbie Caillat', 'Flo Rida',
                'Gucci Mane', 'Young Money', 'Migos', 'Yo Gotti', 'G-Eazy',
                'Foo Fighters', 'Kehlani', 'Princess Nokia', 'French Montana',
                'OneRepublic', 'Elvis', 'Lola Marsh',
                'Lizzo', 'alt-J', 'Hozier', 'The Vaccines',
                'Queen', 'Elton John', 'The Beatles', 'Simon \& Garfunkel',
                'The Lumineers', 'Andy Grammer', 'Gotye', 'Jack Johnson',
                'Amy Winehouse', 'Bob Marley \& The Wailers',
                'AC/DC', 'Noga Erez', 'Noah Kahan', 'Noah Cyrus',
                'John Legend', 'Pessenger', 'Red Hot Chili Peppers',
                'Nirvana', 'Tame Impala', 'Post Malone',
                'Tones And I', 'Vance Joy', 'Ed Sheeran', 'MIKA',
                'The Chainsmokers', 'America', 'ABBA', 'Leon Bridges',
                'The Kooks', 'Arctic Monkeys', 'Eric Clapton', 'theAngelcy',
                'U2', 'slenderbodies', 'Asaf Avidan', 'Augustine',
                'Regina Spektor', 'Stevie Wonder', 'Coldplay',
                'Selena Gomez', 'George Ezra', 'Led Zeppelin', 'Pink Floyd',
                'The Rolling Stones', 'Eagles', 'Aretha Franklin',
                'David Bowie', 'Bee Gees', 'Bob Dylan and The Band',
                'Michael Jackson', 'Prince', 'Madonna', 'Guns N’ Roses',
                'Metallica', 'Tracy Chapman', 'Jason Derulo','System of a Down',
                'Nicki Minaj',
                'Britney Spears', 'Lady Gaga', 'Shakira', 'Ludacris','Akon',
                'Dua Lipa', 'The Weeknd','Miley Cyrus', 'Logic',
                'Rachel Platten', 'Sia', 'Muse', 'Maroon 5', 'Christina Perri',
                'Lorde', 'Jennifer Lopez', 'Cardi B',
                'Major Lazer', 'Imagine Dragons', 'Birdy',
                'John Newman', 'Bruno Mars',
                'P!nk', 'Lana Del Rey', 'Gorillaz', 'Mumford \& Sons',
                'Billie Eilish', 'Ariana Grande', 'Megan Thee Stallion',
                'Doja Cat', 'Halsey', 'Rick Astley', 'Shawn Mendes',
                'Justin Bieber', '24kGoldn', 'Katy Perry', 'Iggy Azalea',
                'Matchbox Twenty', 'Uncle Kracker',
                'Rage Against the Machine', 'The Proclaimers',
                'Future', 'AWOLNATION', 'Weezer', 'MGMT',
                'twenty one pilots', 'The Prodigy', 'Missy Elliott', 'Eve',
                'The Presidents of the United States of America', 'Meek Mill',
                'Drake', 'Macklemore', '21 Savage',
                'T-Pain', 'Ellie Goulding', 'Tenacious D', 'Sublime'.}

\end{document}